\documentclass[conference]{IEEEtran}
\ifCLASSINFOpdf
  \usepackage[pdftex]{graphicx}
\else
\fi
\usepackage{amsmath}
\usepackage{algorithm}
\usepackage{algorithmic}
\usepackage{url}

\usepackage{amssymb}
\usepackage{multirow}
\newcommand{\tab}{\hspace*{0.75em}}

\begin{document}
%
\title{Robust Deep Appearance Models}

\author{\IEEEauthorblockN{Kha Gia Quach$ ^{1} $  $^{*}$,
Chi Nhan Duong$ ^{1} $ $^{*}$,
Khoa Luu$^{2}$ and
Tien D. Bui$ ^{1} $}
\IEEEauthorblockA{$^{1}$ Department of Computer Science and Software Engineering, 
Concordia University, Montreal, Quebec, Canada\\ Email: \{k\_q, c\_duon, bui\}@encs.concordia.ca}
\IEEEauthorblockA{$^{2}$ CyLab Biometrics Center and the Department of Electrical and Computer Engineering,\\ Carnegie Mellon University, Pittsburgh, PA, USA \\
Email: kluu@andrew.cmu.edu}
}

\maketitle

\begin{abstract}
This paper presents a novel Robust Deep Appearance Models to learn the non-linear correlation between shape and texture of face images. In this approach, two crucial components of face images, i.e. shape and texture, are represented by Deep Boltzmann Machines and Robust Deep Boltzmann Machines (RDBM), respectively. The RDBM, an alternative form of Robust Boltzmann Machines, can separate corrupted/occluded pixels in the texture modeling to achieve better reconstruction results.
The two models are connected by Restricted Boltzmann Machines at the top layer to jointly learn and capture the variations of both facial shapes and appearances. 
This paper also introduces new fitting algorithms with occlusion awareness through the mask obtained from the RDBM reconstruction.
The proposed approach is evaluated in various applications by using challenging face datasets, i.e. Labeled Face Parts in the Wild (LFPW), Helen, EURECOM and AR databases, to demonstrate its robustness and capabilities. 
\end{abstract}

\IEEEpeerreviewmaketitle

\section{Introduction}

Active Appearance Models (AAMs) \cite{CootesAAM98} have been used successfully in several areas of facial interpretation over the last two decades. 
Given a new face image, the method aims to “describe” that image by synthesizing a new image similar to it as much as possible.
Indeed, AAMs are statistical models of appearance, generated by combining a shape model that represents the facial structure, and a quasi-localized texture model that represents the pattern of pixel intensities, i.e. skin texture, across a facial image patch. 
However, their capability of generalization is limited by the nature of Principal Component Analysis (PCA) used in both shape and texture models. 
Besides, since AAMs naively combine the shape and texture features to represent the facial appearance also by using PCA, it can only reveal the linear relationship between these features. 
There have been numerous improvements and adaptations using, for example, the probabilistic PCA \cite{medina2014BayesianAAM}, nonlinear Deep Boltzmann Machines (DBM) \cite{salakhutdinov2009deep}, etc., to model large and non-linear variations in  shapes and textures.

\begin{figure}[t]
	\begin{center}
	\includegraphics[width=8cm]{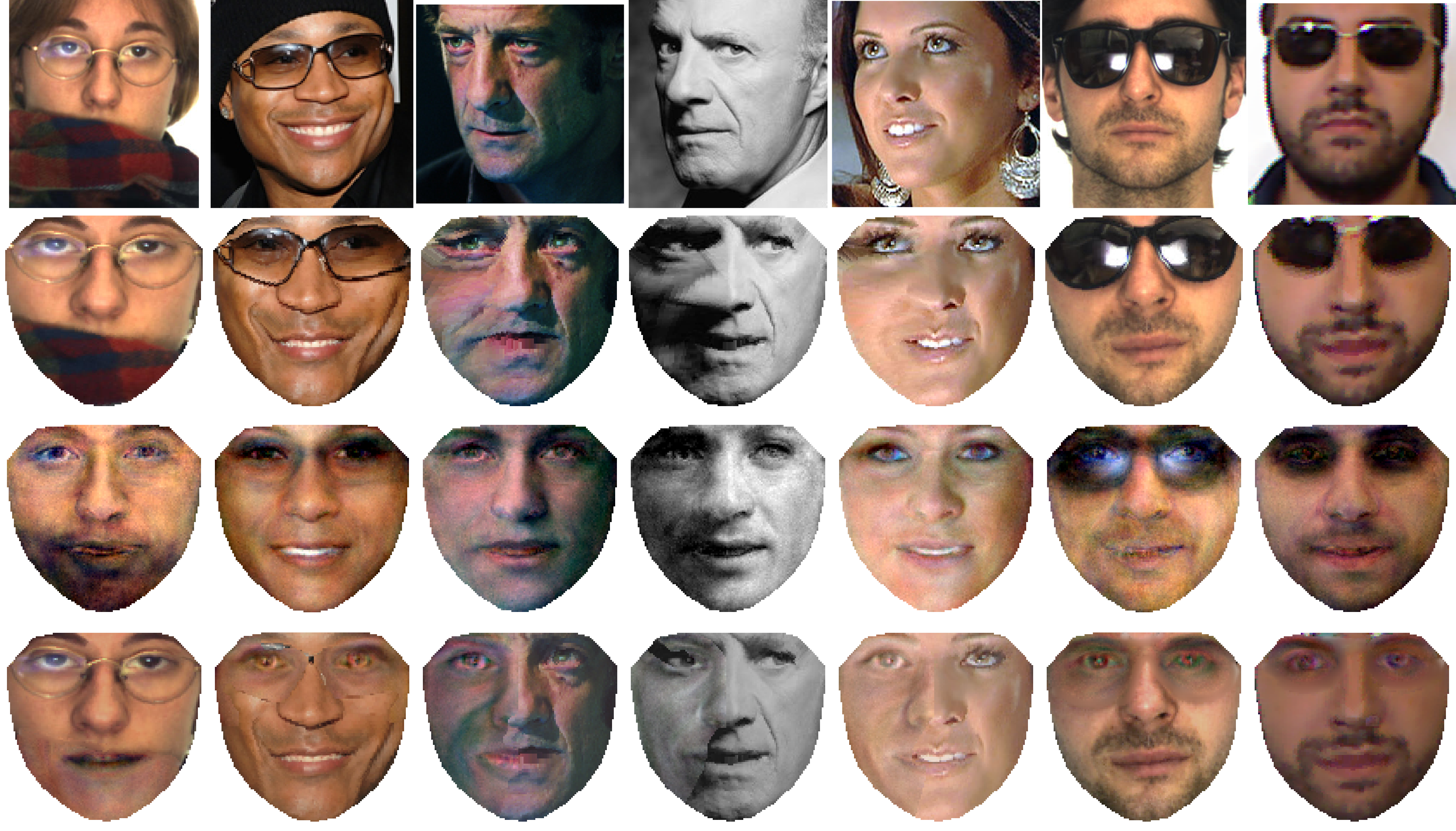}\end{center}
	\caption{Example faces with significant variations, i.e. occlusions and poses, and the modeling results. From top to bottom: original images, shape free images, reconstructed faces using DAMs and our RDAMs approach.}
	\label{fig:ExtremeCaseDAM_known}
\end{figure}
\footnotetext[1]{These two authors contribute equally to the work.}
Duong et al. \cite{Duong_2015_CVPR} recently proposed the Deep Appearance Models (DAMs) approach to model face images using a DBM network.
Their main ideas are first to learn the shape and the texture models of sample faces separately using the DBM approach. The relationships between these two modalities are then pursued to generate the final appearance model using Restricted Boltzmann Machines (RBM) at the top layer. Given an unseen face, DAMs find the optimal facial shape using the forward compositional algorithm. This algorithm minimizes the non-linear least squares error between the warped and the reconstructed textures from the models. This network architecture enables the non-linear modeling capability to overcome the limitations presented in the original AAMs method. 

However, there are still some limitations of DAMs in both face modeling and shape fitting. 
Firstly, the DAMs method still takes into account numerous appearance variations of face images, e.g. facial poses, occlusions, lighting, etc. in their fitting procedure resulting in undesirable fitting performance. 
Their minimization method using the squared error is good enough for constrained face images rather than for the problem of unconstrained face modeling with occlusions, poses and noise. 
Secondly, the texture models of DAMs method cannot distinguish between occluded and non-occluded areas since it treats all regions in the same way during model learning phase. DAMs will capture both ``good" and ``bad" regions in the learned models. Thus, it will give undesirable reconstruction texture images (as shown in Fig. \ref{fig:ExtremeCaseDAM_known}).

To overcome the above modeling and fitting issues, we propose a novel Robust Deep Appearance Models (RDAMs) to learn an additional appearance variation mask that could be used in the fitting procedure to ignore those variations. This mask is modeled by the visible and hidden unit in Robust Boltzmann Machines (RoBM) \cite{tang2012robust}. 
This proposed model not only learns
compact representation for recognition/prediction tasks, but also reconstructs better shape and texture. 

The contributions of this work can be summarized as follows. 
Firstly, we propose a new texture modeling approach named Robust Deep Boltzmann Machines described in section \ref{ssec:RDBM}. It can model  ``good" and ``bad" regions separately via a DBM and a binary RBM, respectively. Then, for example, given a face with sunglasses, RDBM can recover a ``clean" face without sunglasses (as shown in Fig. \ref{fig:RDAMs}). 
Secondly, the proposed RDAMs approach models shape using a DBM since it has non-linear property and can be setup in deep modeling to give more robust representation for shapes. 
Thirdly, we propose to use the learned binary RBM to generate a mask
for shape model fitting using inverse compositional algorithm described in section \ref{ssec:fitting}.
\section{Related Work}

This section reviews Restricted Boltzmann Machines \cite{hinton2002training} and its extensions. Recent advances in AAMs-based facial modeling and fitting approaches are reviewed in this section.

\subsection {RBM and Its Extensions}

Restricted Boltzmann Machines (RBM) \cite{hinton2002training} are an undirected graphical model with two layers of stochastic units, i.e. visible and hidden units, which represent the observed data and the conditional representation of that data, respectively. Visible and hidden units are connected by weighted undirected edges.
Gaussian RBM \cite{krizhevsky2009learning} models real-valued data by assuming the visible units have real values normally distributed with mean $ b_i $ and variance $ \sigma_i^2 $.
Moreover, a set of RBMs can be stacked on top of another to capture more complicated correlations between features in the lower layer. This approach produces a deeper network called Deep Boltzmann Machines \cite{salakhutdinov2009deep}. 
RoBM \cite{tang2012robust} were proposed to estimate noise and learn features simultaneously by distinguishing corrupted and uncorrupted pixels to find optimal latent representations.

\begin{figure}[!t]
	\begin{center}
		\includegraphics[width=8.25cm]{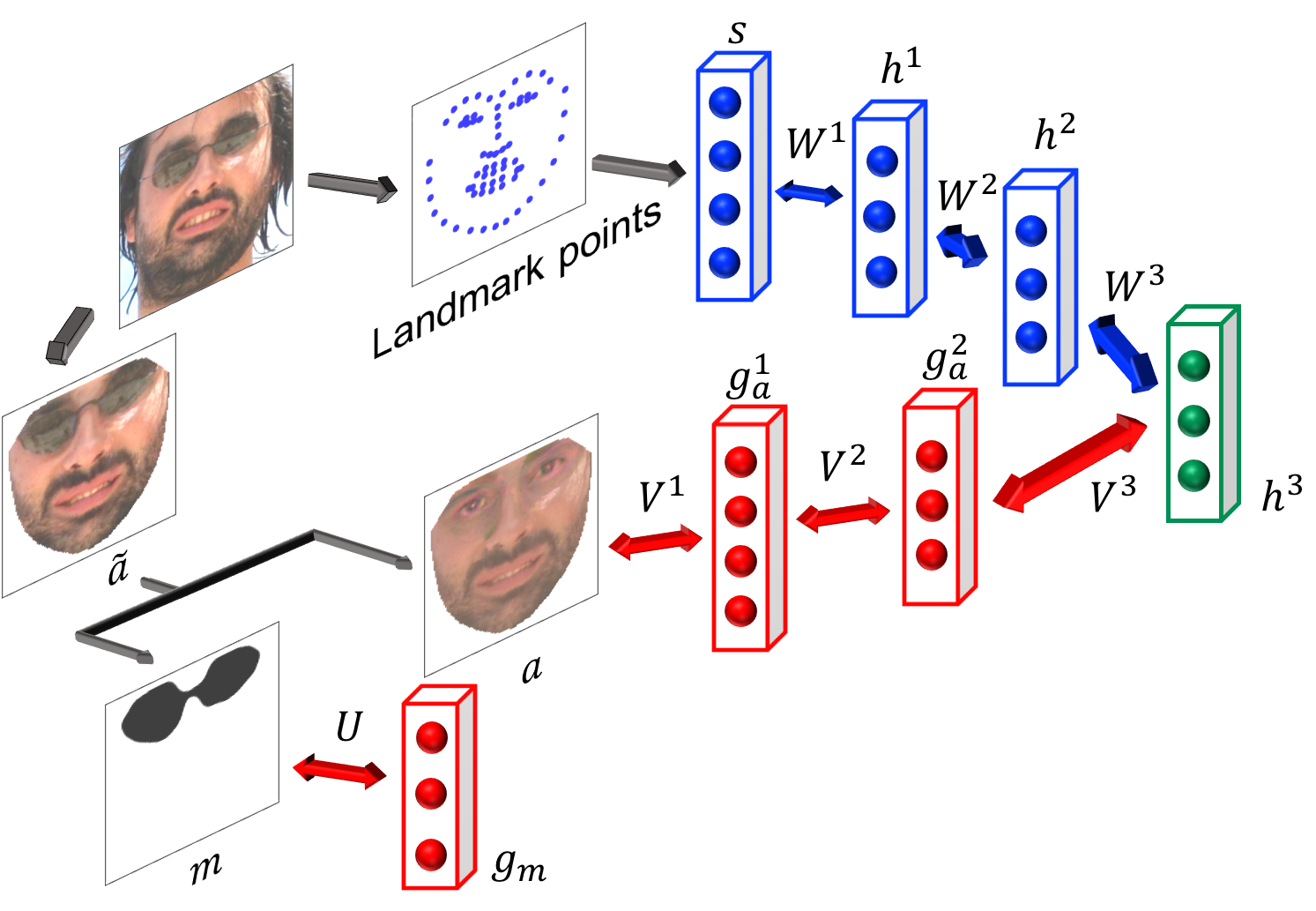}\end{center}
	\caption{The diagram of our RDAMs approach. The blue layers present the shape model with a visible layer $ \mathbf{s} $ and two hidden layers $ \textbf{h}^1$ and $ \textbf{h}^2 $. The red layers denote the texture model with three visible units $ \mathbf{\tilde{a}} $, $ \textbf{a} $ and $ \textbf{m} $, and three hidden layers $ \textbf{g}_m $, $ \textbf{g}^1_a$ and $ \textbf{g}^2_a $. The green layer denotes the appearance model consisting of a hidden layer $ \textbf{h}^3 $}
	\label{fig:RDAMs}
\end{figure}

\subsection {AAMs-based Fitting Approaches}
The fitting steps in AAMs can be formulated as an image alignment problem iteratively solved using the Gaussian-Newton (GN) optimization.
Mathews et al. \cite{matthews2004active} presented the Project Out Inverse Compositional (POIC) algorithm that runs very fast due to pre-computation of the Jacobian and the Hessian matrices. Subsequently, many variants of the IC algorithm have been proposed \cite{baker2003lucas}.
Gross et al. \cite{gross2005generic} introduced the Simultaneous Inverse Compositional (SIC) algorithm simultaneously updating the warp and the texture parameters.
Tzimiropoulos et al. \cite{tzimiropoulos2013optimization} presented the Fast-SIC and the Fast-forward algorithms to efficiently solve the AAMs fitting problem in both forward and inverse fashions. 
An alternative formulation of model fitting is to solve as a classification problem (i.e. distinguish correct and incorrect alignment) or a regression problem. Along this direction, 
Liu \cite{liu2007generic} \cite{liu2009discriminative} proposed to extend GentleBoost classifier for learning discrimination between correct and incorrect alignment; and modeling the nonlinear relationship between texture and parameter updates. 

Due to the holistic nature, AAMs methods are still far from achieving good performance in face images in the wild conditions, e.g. partial occlusions, poses, illumination, etc.
To handle these problems, Sung et al.  \cite{sung2007unified} combined Active Shape Models (ASM) with the AAMs to give a united objective function since ASM can find correct landmark points based on local texture descriptors. 
Tzimiropoulos et al. \cite{tzimiropoulos2011robust} proposed to solve a robust and efficient objective function aiming to detect points under occlusion and illumination changes. 
Martins et al. \cite{martins2013generative} presented two robust fitting methods based on the Lucas-Kanade forwards additive method \cite{lucas1981iterative} to handle partial and self-occlusions.
Recently, Antonakos et al. \cite{antonakos2015active} introduced a graph-based model, called Active Pictorial Structures (APS). This model uses Gaussian Markov Random Field (GMRF) to model the appearance of the objects.
Antonakos et al. \cite{antonakos2015feature} also proposed to use higher level features in face modeling and fitting instead of modeling the raw pixels. 

\section{The Proposed Robust Deep Appearance Models}

This section presents our proposed RDAMs method. 
The structure of RDAMs consists of three main components, i.e. the shape model, the texture model and the appearance representation layer.
Section \ref{ssec:shape_DBM} presents the shape modeling steps using DBM. The robust texture modeling using RDBM is introduced in section \ref{ssec:RDBM}.
Finally, our proposed robust fitting algorithms are presented in section \ref{ssec:fitting}.
The schematic diagram of our proposed method is given in Fig. \ref{fig:RDAMs}.

\subsection{Deep Boltzmann Machines for Shape Modeling}
\label{ssec:shape_DBM}

An $ n $-point shape $ { \mathbf{s} = \left[ x_1, y_1, \cdots, x_n, y_n \right] ^T }$ is modeled using a DBM with a visible layer and two hidden layers. 
Given a shape $\mathbf{s}$, the energy of the configuration $\{\mathbf{s},\mathbf{h}^1,\mathbf{h}^2\}$ of the corresponding layers in shape modeling is as follows,
\begin{eqnarray} \label{eq:ShapeDBM}
\small
\begin{aligned}
E_{DBM_s}(\mathbf{s},\mathbf{h}^1,\mathbf{h}^2;\theta_s)=& \frac{1}{2} \sum_{i}{\frac{(s_i-b_{s_i})^2}{\sigma^2_{si}}}-\sum_{i,j}{W^1_{ij} s_i {h}^1_j}\\
&-\sum_{j,l}{W^2_{jl} {h}^1_j {h}^2_l}
\end{aligned}
\end{eqnarray}
\normalsize
where $ \theta_s = \{\textbf{W}^1, \textbf{W}^2, \sigma_s, \textbf{b}_s\}  $ are the shape model parameters.  
The bias terms of hidden units in two layers in Eqn. \eqref{eq:ShapeDBM} are ignored to simplify the equation. The probability distribution of the configuration $\{\mathbf{s},\mathbf{h}^1,\mathbf{h}^2\}$ is computed as:
\begin{equation}
\small
P(\mathbf{s};\theta_s) =\sum_{\mathbf{h}^1,\mathbf{h}^2} \frac{{\exp\left(-E_{DBM_s} \left( \mathbf{s},\mathbf{h}^1,\mathbf{h}^2;\theta_s \right) \right)  }}{Z(\theta_s)}
\end{equation}
\normalsize
where 
$Z(\theta_s) $ is the normalization constant. 
This shape model is pre-trained using one-step contrastive divergence (CD). 
\subsection{Robust Deep Boltzmann Machines for Texture Modeling}
\label{ssec:RDBM}

We propose a new texture model approach named Robust Deep Boltzmann Machines. Far apart from the texture model of DAMs, this model consists of a visible layer with three gating components: $ \mathbf{a} $,  $ \mathbf{\tilde{a}} $, and $ \textbf{m} $, a binary RBM for the mask variable $ \textbf{m} $ and a Gaussian DBM with the real-valued input variable $ \mathbf{a} $. 
The motivation for using this gating term is to improve modeling and fitting of the DAMs by eliminating the effects of missing, occluded or corrupted pixels.
Our approach uses a Gaussian DBM to model ``clean" data $ \mathbf{a} $ instead of using one Gaussian RBM. There are good reasons for using DBM here. Firstly, it can efficiently capture variations and structures in the input data. Secondly, DBM can deal with ambiguous inputs more robustly due to its top-down feedback.

\subsubsection{Texture Modeling}

Given a shape-free image $\mathbf{\tilde{a}}$, the energy function of the configuration $\{\mathbf{a}, \mathbf{\tilde{a}}, \textbf{m}, \mathbf{g}_m, \mathbf{g}^1_a,\mathbf{g}^2_a\}$ in facial texture modeling is optimized as follows:
\begin{eqnarray}
\small
\label{eq:roDBM}
\begin{aligned}
& E_{RDBM_g} ( \mathbf{a}, \mathbf{\tilde{a}}, \textbf{m},  \mathbf{g}_m,  \mathbf{g}^1_a, \mathbf{g}^2_a; \theta_a ) =
\sum_{i}{\frac{ \gamma_i^2 m_i (a_i -\tilde{a}_i)^2 }{2\sigma^2_{g_i}} } \\
& + \sum_{i}{\frac{(a_i-b_{g_i})^2}{2\sigma^2_{g_i} }}-\sum_{i,j}{ V^{1}_{ij} a_i g^{1}_{aj}} -\sum_{j,l}{V^{2}_{jl} g^{1}_{aj} g^{2}_{al}}\\
& - \sum_{i,k}{ U_{ik} m_i g_{mk}}
+ \sum_{i}{\frac{(\tilde{a}_i - \tilde{b}_{g_i})^2}{2 \tilde{\sigma}^2_{g_i}} }
\end{aligned}
\end{eqnarray}
where $ \theta_a = \{\textbf{V}^1, \textbf{V}^2, \textbf{U}, \sigma_g, \textbf{b}_g, \tilde{\sigma}_g, \mathbf{\tilde{b}}_g\}  $ are the texture model parameters.
It is noted that all the bias terms in Eqn. \eqref{eq:roDBM} are ignored for simplicity. 
The probability distribution of the configuration $\{\mathbf{a}, \mathbf{\tilde{a}}, \textbf{m}, \mathbf{g}_m, \mathbf{g}^1_a,\mathbf{g}^2_a\}$ is computed as follow:
\begin{equation}
\small
P(\mathbf{\tilde{a}};\theta_a)
=\sum_{\mathbf{g}^1_a,\mathbf{g}^2_a}{ \frac{\exp \left( -E_{RDBM_g} \left(\mathbf{a},\mathbf{\tilde{a}}, \textbf{m}, \mathbf{g}_m, \mathbf{g}^1_a,\mathbf{g}^2_a;\theta_a \right) \right) }{Z(\theta_a)}}
\end{equation}
Given the input variables $\mathbf{\tilde{a}}$, the states of all layers can be inferred by computing the posterior probability of the latent variables, i.e. $ p ( \mathbf{a}, \textbf{m},  \mathbf{g}_m,  \mathbf{g}^1_a, \mathbf{g}^2_a | \mathbf{\tilde{a}} ) $. Therefore, the sampling can be divided into two folds, i.e. one for the visible units and one for the hidden units. For the visible variables $ \mathbf{a} $ and $ \textbf{m} $, the conditional distributions can be sampled as, 
\begin{eqnarray}
\small
p( \mathbf{a}, \textbf{m} |  \mathbf{g}_m,  \mathbf{g}^1_a, \mathbf{\tilde{a}} ) = p( \mathbf{a} |  \textbf{m}, \mathbf{g}^1_a, \mathbf{\tilde{a}} ) p( \textbf{m} |  \mathbf{g}_m,  \mathbf{g}^1_a, \mathbf{\tilde{a}} ) 
\end{eqnarray}
For the hidden variables $ \mathbf{g}_m,  \mathbf{g}^1_a, \mathbf{g}^2_a  $, the conditional distributions can be sampled as follows, 
\begin{equation}
\small
p ( \mathbf{g}_m,  \mathbf{g}^1_a, \mathbf{g}^2_a | \mathbf{a}, \textbf{m}, \mathbf{\tilde{a}} ) = p( \mathbf{g}_m | \textbf{m} ) p ( \mathbf{g}^1_a |  \mathbf{a}, \mathbf{g}^2_a ) p(\mathbf{g}^2_a | \mathbf{g}^1_a) 
\end{equation}
The sampling process can be applied on each unit separately since the distribution is factorial.  
Section \ref{sssec:TextureML} will discuss the learning procedure of this texture model.

\subsubsection{Model Learning for RDBM}
\label{sssec:TextureML}

To pre-train our presented RDBM model, the DBM, which models ``clean" faces, is first trained with some ``clean" images and then the parameters in the RDBM model are optimized to maximize the log likelihood as follows,
\begin{equation} \label{eq:obj_robm}
\small
\theta_a^* = \arg \max_{\theta_a} \log{P(\mathbf{\tilde{a}};\theta_a)}
\end{equation}
The optimal parameter values can then be obtained using a gradient descent procedure given by,
\begin{equation}
\small
\frac{\partial}{\partial \theta_a}\mathbb{E}\left[\log P(\mathbf{\tilde{a}};\theta_a) \right]=\mathbb{E}_{P_{\text{data}}}\left[\frac{\partial E_{\text{RDBM}_g}}{\partial \theta_a}\right]-\mathbb{E}_{P_{\text{model}}}\left[\frac{\partial E_{\text{RDBM}_g}}{\partial \theta_a}\right]
\end{equation}
where $\mathbb{E}_{P_{\text{data}}}\left[ \cdot \right]$ and $\mathbb{E}_{P_{\text{model}}}\left[ \cdot \right]$ are the expectations respecting to data distribution and distribution estimated by the RDBM.
The two terms can be approximated using mean-field inference and Markov Chain Monte Carlo (MCMC) based stochastic approximation, respectively. 

\begin{figure}[t]
	\begin{center}
		\includegraphics[height=2.7cm]{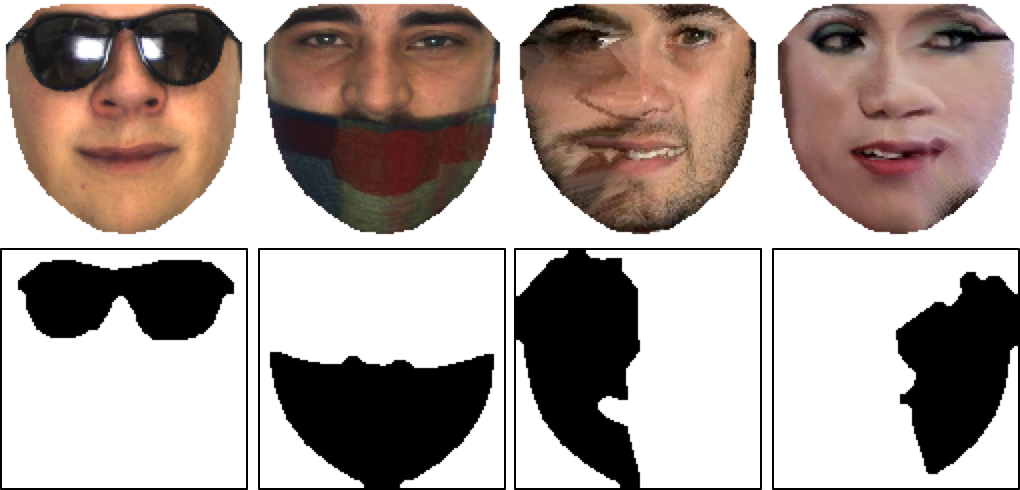}\end{center}
	\caption{Examples of automatically detected masks from the shape-free images. Top row: shape-free images. Bottom row: detected binary masks using the technique in section \ref{sssec:TextureMaskRBM}}
	\label{fig:Mask_RBM}
\end{figure}

In our method, pre-training the parameters of the DBM on ``clean" data first will make the process of learning the texture model faster and much easier. Similarly, we also propose to first learn the parameters of the binary RBM (to represent the mask $ \textbf{m} $) on pre-defined and extracted training masks (as shown in Fig. \ref{fig:Mask_RBM}) instead of randomizing the parameters. Then, we propose an automatic technique to extract such training masks for learning the binary RBM in the next section. 

\subsubsection{Learning Binary Mask RBM}
\label{sssec:TextureMaskRBM}

This section aims to generate masks from the training images having poses and occlusions, e.g. sunglasses and scarves. We consider learning three types of binary mask, i.e. sunglasses, scarves and pose stretching. A binary RBM is learned to represent each type of mask. 
Binary masks for sunglasses or scarves can be extracted by applying a global threshold on shape-free images having sunglasses or scarves with a prior knowledge of their locations.
We will focus on the last and hardest type, i.e. pose stretching. 

\begin{figure}[!t]
	\begin{center}
		\includegraphics[width=6.5cm]{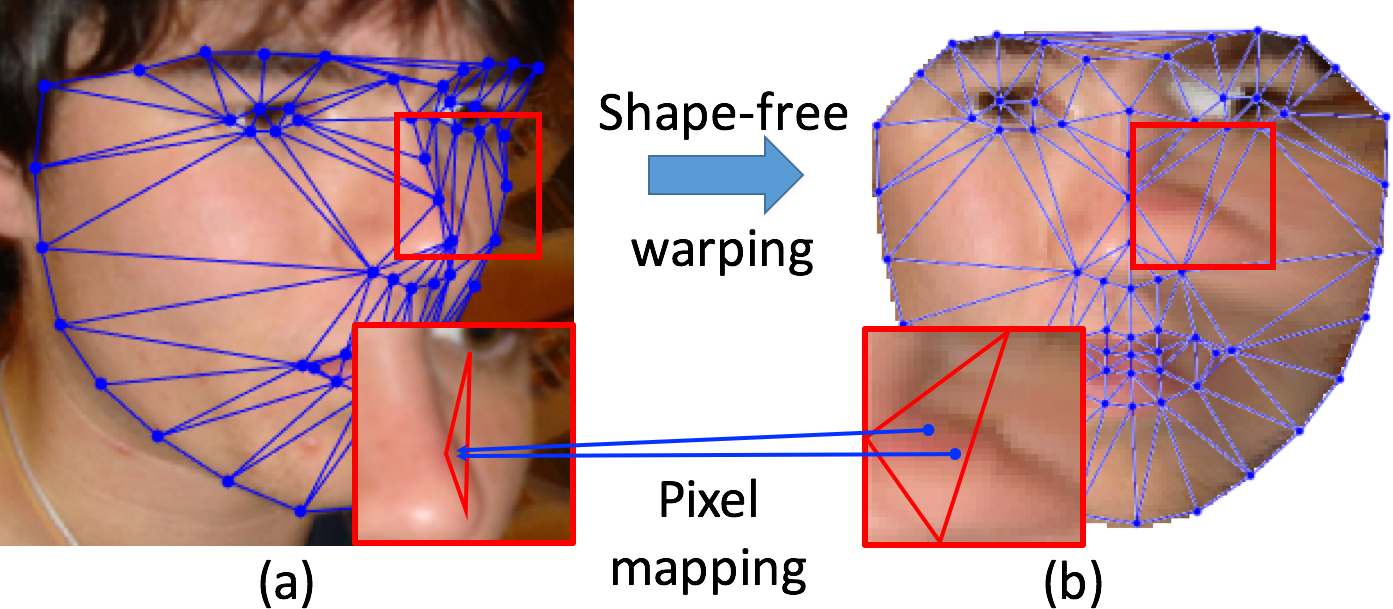}\end{center}
	\caption{An illustration in pose stretching detection: (a) Source image (b) Target warped shape-free image}
	\label{fig:Pose_stretch}
\end{figure}

In 2D texture model, warping faces with a large pose (e.g. larger than $ \pm 45^\circ $) will likely cause stretching effects on half of the faces since the same pixel values are copied over a large region (see Fig. \ref{fig:Pose_stretch}).
Therefore, we propose a technique that can detect such stretching regions during warping process. 
The main idea is to count the number of unique pixels in the source triangle that are mapped to the pixels in the target triangle. As we know, a source pixel can be mapped to multiple target pixels due to interpolation. The degree of a target triangle being stretched is equivalent to $ 	p = (\frac{n_0}{N})  $, where $ p = 1 $ means there is no stretching and the stretching is visible when $ p < 0.9 $ (as from our experiments), $ n_0 $ and $ N $ are the number of unique pixels and the total number of pixels in the corresponding source triangle, respectively.
Finally, we can use the detected regions as a mask to pre-train the above robust texture model.

\subsection{Model Fitting Algorithms in RDAMs}
\label{ssec:fitting}

With the trained shape and texture models, the process of finding an optimal shape of a new image $ I $ can be formulated as finding an optimal shape $\mathbf{s}$ that maximizes the probability of the shape-free image as 
${\mathbf{s}^*=\arg\max_{\mathbf{s}}P(I(\textbf{W}(r_{\mathcal{D}},\mathbf{s}))|\mathbf{s};\theta)} $.

During the fitting steps, the states of hidden units $\mathbf{g}^1_a$ are estimated by clamping both the current shape $\mathbf{s}$ and the warped texture $\mathbf{\tilde{a}}$ to the model. The Gibbs sampling method is then applied to find the optimal estimated ``clean" texture $ \mathbf{a} $ of the testing face given the current shape $\mathbf{s}$. 
Let $\mathbf{a}=\sigma_g \mathbf{V}^{1} \mathbf{g}^1_a+\mathbf{b}_g $ be the mean of the Gaussian distribution, we have
$P(I(W(r_{\mathcal{D}},\mathbf{s}))|\mathbf{g}^1_a;\theta)=\mathcal{N}(\mathbf{a},\sigma_g^2\mathbf{I})$
where $ \textbf{I} $ is the identity matrix. The maximum likelihood can then be estimated as
${\mathbf{s}^*=\arg\max_\mathbf{s}\mathcal{N}(I(\textbf{W}(r_{\mathcal{D}},\mathbf{s}))|\mathbf{a},\sigma_g^2\mathbf{I})}=\arg\min_\mathbf{s}\frac{1}{\sigma_g^2} \| I(\textbf{W}(r_{\mathcal{D}},\mathbf{s})) -\mathbf{a} \| ^2 $.

This brings us to the non-linear least squares problem solved in image alignment.
Notice that $ \textbf{a} $ is the reconstructed ``clean" texture while $ I(W(r_{\mathcal{D}},\mathbf{s})) $ is the warped texture from the input image. If the input image contains occlusion or corruption, it is clearly that the above square error will not reflect the goodness of the current shape $ \mathbf{s} $. Thus, solely using $ \ell_2 $-norm may limit the performance of shape fitting and reconstruction of the models. 
Since our proposed model can generate a mask of corrupted pixels, we propose to incorporate the mask $ \mathbf{m}$ into the original objective function as:
\begin{equation}\label{eq:NLS_fitting}
\small
\mathbf{s}^* =\arg\min_\mathbf{s}  \|  \mathbf{m} \odot \left( I(\textbf{W}(r_{\mathcal{D}},\mathbf{s})) -\mathbf{a} \right)\| ^2
\end{equation}
where $ \odot $ is the component-wise multiplication. 

\begin{algorithm}[!t]
	\small
	\caption{\textbf{$ - $ Inverse Compositional}}
	\label{alg:inverse_compositional}
	\begin{algorithmic}
		\STATE 1. {\bfseries Pre-compute:} The gradient $ \nabla \mathbf{a}  $, the Jacobian $ \frac{\partial \textbf{W}}{ \partial \mathbf{s} } $ at $(r_{\mathcal{D}}; 0) $, the steepest descent $ SD = \nabla \mathbf{a} \frac{\partial \textbf{W}}{ \partial \mathbf{s}} $, the Hessian matrix $ H = SD^T SD $\\
		\STATE 2. {\bfseries At each iteration:} \\
		\tab \tab (I) Perform warping operator $ \textbf{W} $ to obtain warped texture $ I(\textbf{W}(r_{\mathcal{D}},\mathbf{s})) $\\
		\tab \tab (II) Compute the texture reconstruction error $  {( \textbf{m} \odot (I(\textbf{W}(r_{\mathcal{D}},\mathbf{s})) - \mathbf{a}))}$\\
		\tab \tab (III)	Compute $ \nabla \mathbf{a} \frac{\partial \textbf{W}}{ \partial \mathbf{s}}  (\textbf{m} \odot (I(\textbf{W}(r_{\mathcal{D}},\mathbf{s})) - \mathbf{a}) ) $\\
		\tab \tab (IV) Compute $ \Delta \mathbf{s} $ using Eqn. \eqref{eq:update_s_inverse_compositional} \\
		\tab \tab (V) Update the shape parameters by composing the warp operator 
		$ \mathbf{s} \rightarrow \mathbf{s} \circ \Delta \mathbf{s}^{-1} $
        \end{algorithmic}
\end{algorithm}

The inverse compositional algorithm tries to minimize the incremental warp computed with respect to the model image $ \mathbf{a} $ instead of with respect to $ I ( \textbf{W}(r_\mathcal{D}, \mathbf{s}))$. 
\begin{equation} \label{eq:update_s_inverse_compositional}
\small
\Delta \mathbf{s} = \arg\min_{\Delta \mathbf{s}} \| \textbf{m} \odot  \left(I(\textbf{W}(r_{\mathcal{D}}, \mathbf{s}))-\mathbf{a} ( \textbf{W} ( r_{\mathcal{D}}, \Delta\mathbf{s} )  ) \right) \|^2
\end{equation}
with respect to $ \Delta\mathbf{s} $ and then updating the parameters as $ { \mathbf{s} \leftarrow \mathbf{s} \circ \Delta \mathbf{s}^{-1} }$, where $ \circ $ denotes the composition of two warps. The solution of the least squares problem above is
${\Delta \mathbf{s} = \textbf{H}^{-1} \textbf{J}^T_{\mathbf{a}} ( \mathbf{m} \odot (I(\textbf{W} ( r_{\mathcal{D}}, \mathbf{s}))  - \mathbf{a}) )}$
where $\mathbf{J}_{\mathbf{a}} = \nabla \mathbf{a} \frac{\partial \textbf{W}}{\partial \mathbf{s}}$ is the Jacobian matrix of the model image $ \mathbf{a} $. The Hessian matrices $ \textbf{H} $ are then given by $ {\textbf{H} = ( \textbf{m} \odot \textbf{J}_{\mathbf{a}})^T (  \textbf{m} \odot \textbf{J}_{\mathbf{a}} )} $.
\begin{figure*}[!t]
	\begin{center}
		\includegraphics[width=15.75cm]{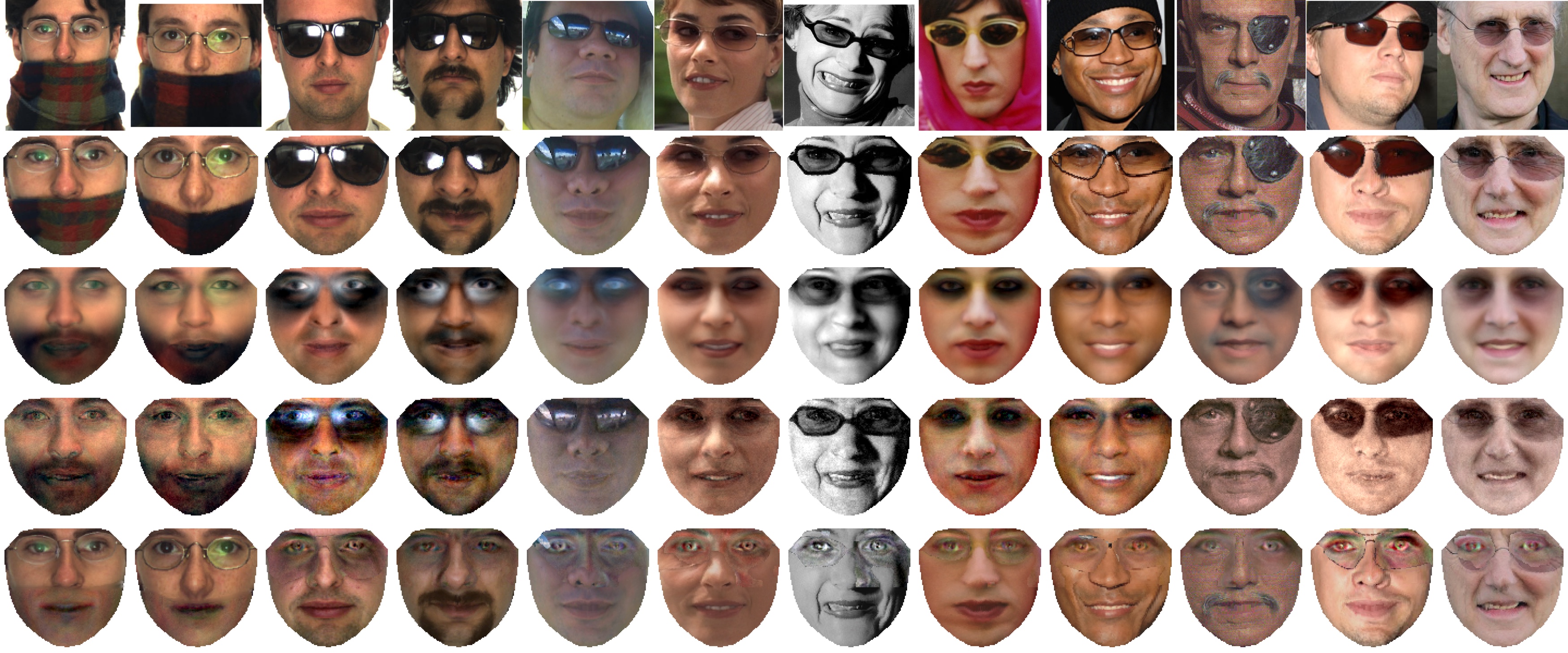}\end{center}
	\caption{Reconstruction results on images with occlusions (i.e. sunglasses or scarves) in LFPW, Helen and AR databases.  The first row: input images, the second row: shape-free images, from the third to fifth rows: reconstructed results using AAMs, DAMs and RDAMs, respectively.}
	\label{fig:reconstruction_lfpw_sung}
\end{figure*}

\section{Experiments}

In this section, we evaluate the performance of our proposed framework in face modeling tasks using data ``in the wild" (sections \ref{ssec:FaceOclRem} and \ref{ssec:exp_pose_recovery}
). Then we demonstrate its robustness in model fitting steps (section \ref{ssec:exp_face_fitting}).

\subsection{Databases} \label{ssec:databases}
\label{subsec:db}

The \textbf{LFPW} \cite{belhumeur2011localizing} database consists of 1400 images but only about 1000 images are available (811 for training and 224 for testing). For each image, we have 68 landmark points provided by 300-W competition \cite{sagonas2013semi}.

The \textbf{Helen} \cite{le2012interactive} database contains about 2300 high-resolution images (2000 for training and 300 for testing). 68 landmark points are annotated for all faces. 
The facial images contain different poses, expressions and occlusions. 

The \textbf{AR} database \cite{martinez1998ar} contains 134 people (75 males and 59 females) and each subject has 26 frontal images (14 normal images with different lighting and expressions, six occluded images with sunglasses and six for scarves).

The \textbf{EURECOM} database \cite{min_2014} consists of facial images of 52 people (38 males and 14 females). Each person has different expressions, lighting and occlusion conditions. We only use images wearing sunglasses in our experiments.

\subsection{Facial Occlusion Removal}
\label{ssec:FaceOclRem}

In this section, we demonstrate the ability of RDAMs to handle extreme cases of occlusions such as sunglasses or scarves. 
RDAMs are trained in two steps: \textit{pre-train each layer} and \textit{train the whole model}.
The training set includes 1000 ``clean" and 200 posed images from \textbf{LFPW} and \textbf{Helen}, 534 ``clean", 95 sunglasses, and 95 scarf images from 95 subjects in \textbf{AR}, 104 images from 52 subjects in \textbf{EURECOM}. 
For \textit{the pre-training steps}, we first train shape DBM using all shapes. 
Then, we train RDBM by first separately training GRBM with clean images 
and learning binary mask RBM with masks generated from occluded and posed images in AR, EURECOM or LFPW. 
After that, we can train the RDBM with pre-initialized weights of GRBM and mask RBM. 
The joint layer is later trained with all training images. Each step above is trained using Contrastive Divergence learning in 600 epochs on a system of Xeon@3.6GHz CPU, 32.00GB RAM.  The computational costs (without parallel processing) are as follows. The \textbf{training time} is \textit{14.2 hours}. \textbf{Fitting} on average is \textit{17.4s}. \textbf{Reconstructing faces} on average is \textit{1.53s}.

\begin{figure}[b]
	\begin{center}
		\includegraphics[width=0.85\linewidth]{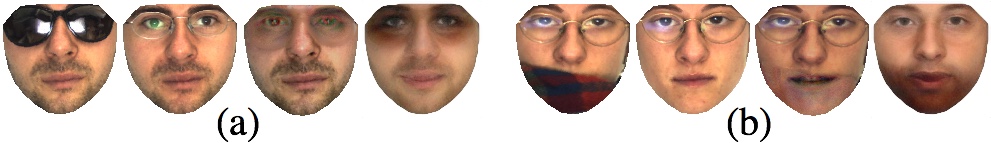}
	\end{center}
	\caption{Reconstruction results on images with sunglasses (a) or scarves (b) in AR database. 
		The images are input shape-free, ground truth shape-free, reconstructed results using RDAMs and RPCA \cite{quach2014sparse}, respectively.
	}
	\label{fig:recon_gt}
\end{figure}

As shown in Fig. \ref{fig:reconstruction_lfpw_sung}, RDAMs can remove those occlusions successfully without leaving any severe artifact comparing with the baseline AAMs method and the state-of-the-art DAMs method. We also compare with RPCA-based method \cite{quach2014sparse} (See Fig. \ref{fig:recon_gt}).
We measure the reconstruction quality in terms of Root Mean Square Error (RMSE) on LFPW, Helen, AR and EURECOM databases in different ways.
\begin{table}[t]
	\small
	\caption{The average RMSEs of reconstructed images using different methods on LFPW and AR databases with sunglasses and scarf}
	\label{tab:RMSE_occlusion_sung_scarf}
	\centering
	\begin{tabular}{|c|c|c|c|}
		\hline
		\textbf{Methods} & AAMs \cite{tzimiropoulos2013optimization}  & DAMs \cite{Duong_2015_CVPR} & \textbf{RDAMs}\\
		\hline
		\textbf{LFPW \& Helen} & 12.91 (18.98) & 11.15 (14.98) & \textbf{8.58} (23.98) \\
		\hline
		\textbf{AR - Sunglasses} & 56.55 & 55.48 & \textbf{41.67}\\
		\hline
		\textbf{AR - Scarf} & 63.16  & 60.96 & \textbf{47.65}  \\
		\hline
	\end{tabular}
\end{table}
\begin{figure}[!b]
	\begin{center}
		\includegraphics[width=8.5cm]{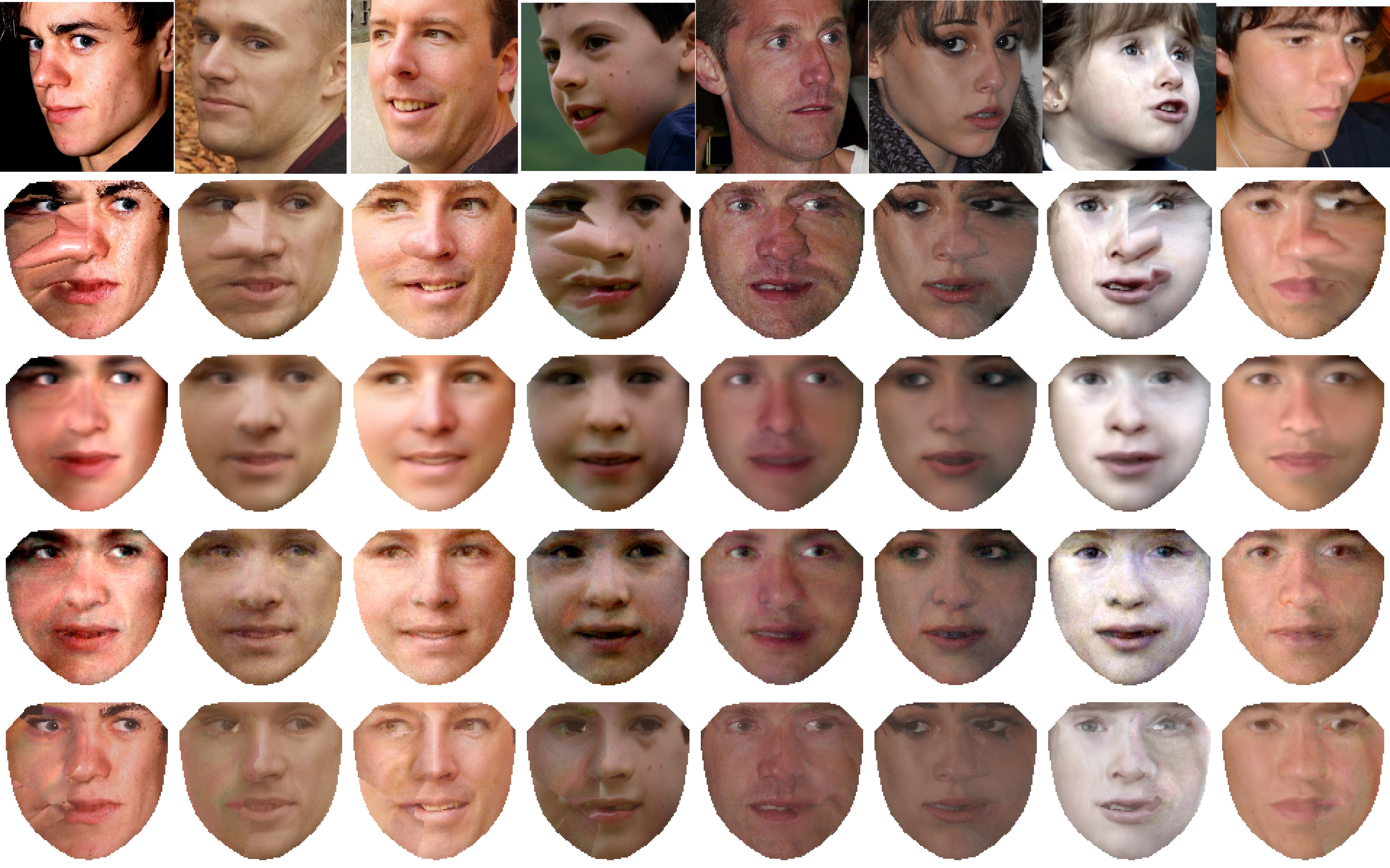}\end{center}
	\caption{Facial pose recovery results on images from LFPW and Helen databases.  The first row is the input images. The second row is the shape-free images. From the third to fifth rows are AAMs, DAMs and RDAMs reconstruction, respectively.}
	\label{fig:exp_headpose_reconstruct}
\end{figure}
We choose from AR two subsets of 210 images with sunglasses and 210 images with scarves from 38 subjects (30 males and eight females) not in the training set. The corresponding normal face images, i.e. frontal and without occlusions, of the same person are used as the references to compute the RMSE.
We select a subset of 23 images with sunglasses and 100 images with some occlusions from LFPW and Helen.  
A mask is used to ignore occluded/corrupted pixels in the testing images so that we have an unbiased metrics. 
The average masked-RMSEs of AAMs, DAMs and our RDAMs are shown in Table \ref{tab:RMSE_occlusion_sung_scarf}. The average unmasked-RMSEs are also reported for reference (i.e. the numbers inside the brackets). 
Our RDAMs achieve the best reconstruction results compared against AAMs and DAMs. 
Note that the unmasked-RMSE is always higher than masked-RMSE since some corrupted pixels are recovered during reconstruction. Since our RDAMs can recover more corrupted pixels, it makes the un-masked RMSE higher than the ones from AAMs and DAMs. 

\begin{figure}[!t]
	\begin{center}
		\includegraphics[width=8.5cm]{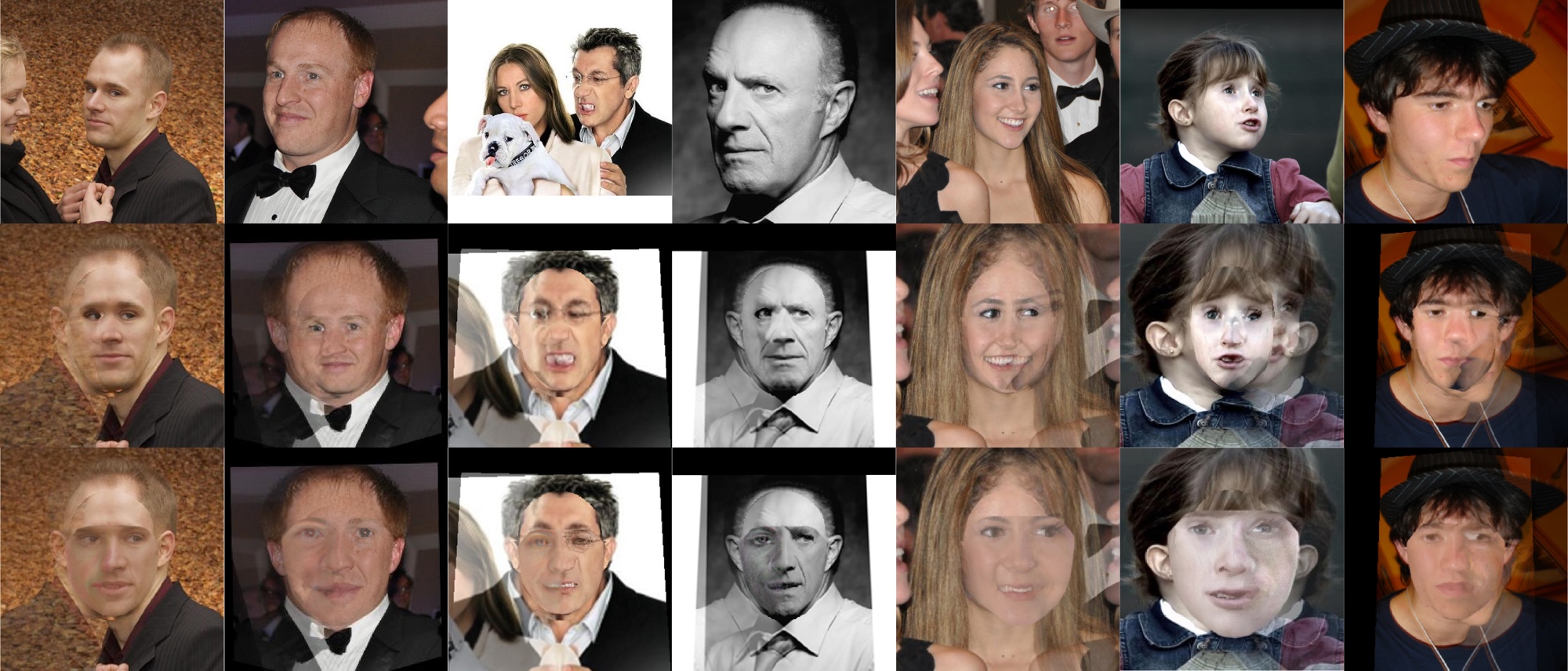}\end{center}
	\caption{Comparisons between RDAMs and Face Frontalization approach \cite{Hassner_2015_CVPR}. The 1st row: input faces; the 2nd row: synthesized frontal view with soft symmetry background \cite{Hassner_2015_CVPR}; the 3rd row: frontalized faces generated by RDAMs on the same background.}
	\label{fig:exp_frontalization_reconstruct}
\end{figure}

\subsection{Facial Pose Recovery}
\label{ssec:exp_pose_recovery}

This section illustrates the capability of RDAMs to deal with facial poses. Using the same pre-trained model presented in Section \ref{ssec:FaceOclRem}, the texture model was trained using 280 images with different pose variations from LFPW and Helen databases. 
The reconstruction results of facial images with different poses are presented in Fig. \ref{fig:exp_headpose_reconstruct}. 
In this experiment, our RDAMs also achieve the best reconstruction results comparing to AAMs and DAMs especially in the cases of extreme poses (more than $ 45^\circ $). Our proposed RDAMs method can handle those extreme poses in a more natural way. From Fig. \ref{fig:exp_headpose_reconstruct}, RDAMs give reconstructed faces that look more similar to the original faces while DAMs or AAMs make the face look younger or change its identity.

Another experiment is performed to demonstrate our RDAMs approach on the face frontalization problem. Given an input face with poses, the process of “frontalization” is to synthesize the frontal view of that face. Our RDAMs approach is compared with the state-of-the-art frontalization method \cite{Hassner_2015_CVPR} on LFPW and Helen databases as shown in Fig. \ref{fig:exp_frontalization_reconstruct}. RDAMs only model certain facial areas not including hair, forehead, neck and ears. For the ease of comparison, the reconstructed texture of RDAMs (the last row) is put on top of the corresponding images in the middle row.  
Although we lost some color consistency with the background, RDAMs can produce more natural looking faces than images in \cite{Hassner_2015_CVPR}.

\begin{table}[t!]
	\small
	\caption{The average MSE between estimated shape and ground truth shape (68 landmark points).}
	\label{tab:RMSE_mask_nomask}
	\centering
	\begin{tabular}{|c|c|c|c|c|}
		\hline
		\textbf{Method} & \textbf{Initial} & \textbf{RDAMs} & \textbf{Fast-SIC} \cite{tzimiropoulos2013optimization} & \textbf{AOMs}  \cite{tzimiropoulos2014active} \\
		\hline
		Sunglasses & 0.195 & 0.1672 & \textbf{0.1218} & 0.1705 \\
		\hline
		Scarves & 0.211 & \textbf{0.0756} & 0.0756 & 0.1705 \\
		\hline
	\end{tabular}
\end{table}
\subsection{Model Fitting in RDAMs} \label{ssec:exp_face_fitting}
We compared our results with Active Orientation Models \cite{tzimiropoulos2014active} and Fast-SIC \cite{tzimiropoulos2013optimization} in the following modeling fitting experiment.
We evaluated model fitting using the LFPW and the AR databases with about 300 images (23 images from LFPW database, and 268 images from AR database. The average errors are reported in Table \ref{tab:RMSE_mask_nomask}.  The initial shape is the mean shape placed inside the face’s bounding box. RDAMs achieve comparable performance compared to other methods.

\section{Conclusion}

In this paper, the novel Robust Deep Appearance Models have been proposed to deal with large variations in the wild such as occlusions and poses.
Comparing with the previous DAMs model, the proposed approach can produce remarkable reconstruction results even when faces are occluded or having extreme poses.
Moreover, the proposed fitting algorithms fit well with the new texture model such that it can make use of the occlusion mask generated by the proposed model.
Experimental results in occlusion removal, pose correction and model fitting have shown the robustness of the model against large occlusions and poses.
\section*{Acknowledgment}
This work is supported in part by the Natural Sciences and Engineering Research Council (NSERC) of Canada
{\small
	\bibliographystyle{IEEEtran}
	\bibliography{egbib}
}

\end{document}